\documentclass[runningheads,a4paper]{llncs}
\usepackage{wrapfig}
\usepackage{graphicx}
\usepackage{booktabs}
\usepackage{colortbl}
\usepackage{subcaption}
\usepackage{hyperref}
\hypersetup{
hidelinks
}
\usepackage{amsmath}
\usepackage{multirow}
\usepackage{enumitem}
\usepackage{cite}
\usepackage{amssymb}
\usepackage{graphicx}
\usepackage{amsmath}
\usepackage{url}
\usepackage{algorithm}
\usepackage{algorithmic}
\usepackage{marvosym}
\usepackage[table]{xcolor} 
\definecolor{Gray}{gray}{0.9}

\usepackage{bbding}
\newcommand{\keywords}[1]{\par\addvspace\baselineskip
\noindent\keywordname\enspace\ignorespaces#1}

\begin{document}

\mainmatter

\title{AU-LLM: Micro-Expression Action Unit Detection via Enhanced LLM-Based Feature Fusion}

\titlerunning{AU-LLM: Enhancing Micro-Expression AU Detection with LLMs}

%
%
\author{Zhishu Liu\textsuperscript{1}\thanks{Equal contribution. \quad \textsuperscript{(\Letter)} Zitong Yu is the corresponding author. } \and Kaishen Yuan\textsuperscript{1}\footnotemark[1] \and Bo Zhao\textsuperscript{1}\and Yong Xu\textsuperscript{2} \and Zitong Yu\textsuperscript{1}\textsuperscript{(\Letter)}}

%
\authorrunning{Zhishu Liu et al.} 
\institute{\textsuperscript{1}Great Bay University \\
\textsuperscript{2}Harbin Institute of Technology , Shenzhen}
%
\toctitle{Lecture Notes in Computer Science}
\tocauthor{Authors' Instructions}

\maketitle

\begin{abstract} 
The detection of micro-expression Action Units (AUs) is a formidable challenge in affective computing, pivotal for decoding subtle, involuntary human emotions. While Large Language Models (LLMs) demonstrate profound reasoning abilities, their application to the fine-grained, low-intensity domain of micro-expression AU detection remains unexplored. This paper pioneers this direction by introducing \textbf{AU-LLM}, a novel framework that for the first time uses LLM to detect AUs in micro-expression datasets with subtle intensities and the scarcity of data. We specifically address the critical vision-language semantic gap, the \textbf{Enhanced Fusion Projector (EFP)}. The EFP employs a Multi-Layer Perceptron (MLP) to intelligently fuse mid-level (local texture) and high-level (global semantics) visual features from a specialized 3D-CNN backbone into a single, information-dense token. This compact representation effectively empowers the LLM to perform nuanced reasoning over subtle facial muscle movements.Through extensive evaluations on the benchmark CASME II and SAMM datasets, including stringent Leave-One-Subject-Out (LOSO) and cross-domain protocols, AU-LLM establishes a new state-of-the-art, validating the significant potential and robustness of LLM-based reasoning for micro-expression analysis. The codes are available
at \href{https://github.com/ZS-liu-JLU/AU-LLMs}{\textcolor{blue}{Link}}.

\end{abstract}

\vspace{-2.8em}
\section{Introduction}
\vspace{-0.5em}
Facial Action Units (AUs), defined by the Facial Action Coding System (FACS)~\cite{ref:facs}, are the atomic components of facial expressions, making their detection fundamental to affective computing. Unlike overt macro-expressions, micro-expressions are brief, low-intensity, involuntary muscle twitches that betray concealed emotions. Their fleeting nature creates an extremely low signal-to-noise ratio, demanding exceptional model sensitivity to capture faint visual cues. While traditional deep learning models like CNNs and Transformers have advanced AU detection~\cite{ref:au_cnn, ref:au_transformer, ref:AUformer}, they often lack the capacity to reason about the complex interplay of AUs in such subtle conditions.

The ascendancy of Large Language Models (LLMs) has established a new paradigm for complex reasoning~\cite{ref:gpt3, ref:flamingo}, offering a promising avenue for interpreting the 'grammar' of facial movements. However, applying this potential to micro-expression AU detection is challenged by the semantic gap between continuous visual features and the LLM's discrete token space~\cite{ref:li2019semantic}. A naive projection risks losing the critical low-intensity signals, raising the question: how can we distill the essence of a fleeting facial movement into a representation an LLM can effectively reason about?

To bridge this gap, we introduce \textbf{AU-LLM}, the first framework to leverage LLM reasoning for micro-expression AU detection. Our approach features the \textbf{Enhanced Fusion Projector (EFP)}, an MLP-based module designed to fuse crucial mid-level (local texture) and high-level (global context) visual features into a single, information-dense token. This provides the LLM with a rich, distilled representation for nuanced reasoning, and we adapt the model efficiently using Low-Rank Adaptation (LoRA)~\cite{ref:lora}. Extensive experiments from both within-domain (LOSO) and cross-domain perspectives demonstrate that AU-LLM significantly outperforms state-of-the-art methods, showcasing its powerful generalization and reasoning capabilities. Our contributions can be summarized as follows:
\begin{itemize}[leftmargin=*]
    \item We propose AU-LLM, the first framework to successfully apply LLM-based reasoning to the challenging task of micro-expression AU detection.
    \item We introduce the EFP module, an effective method for fusing multi-level visual features into a compact token to combine crucial local textural details with global semantic context, thereby bridging the vision-language gap more effectively.
    \item Our model outperforms state-of-the-art methods on the CASME II and SAMM datasets, validated through comprehensive experiments from both within-domain (LOSO) and cross-domain perspectives.
\end{itemize}

\vspace{-0.75em}
\section{Related Work}
\vspace{-0.75em}

\noindent\textbf{Micro-Expression AU Detection.}
Automated Facial Action Unit (AU) detection is a core task in affective computing~\cite{ref:xie2020assisted}. While detecting AUs in macro-expressions is well-established, micro-expression AU detection is significantly more challenging due to their fleeting, low-intensity, and involuntary nature~\cite{ref:ttt}. To overcome this, recent works enhance feature learning by either improving the visual signal itself (e.g., LED~\cite{ref:led}, InfuseNet~\cite{ref:infusenet}) or learning more discriminative representations via attention (e.g., SCA~\cite{ref:sca}) and advanced training strategies like knowledge distillation (e.g., DVASP~\cite{ref:dvasp}) and contrastive learning (e.g., IICL~\cite{ref:iicl}). Despite these advances, their primary focus is on refining the visual representation for a simple final classifier, limiting high-level semantic reasoning. Their focus remains on what features to extract, not how to reason about them. Our work, \textbf{AU-LLM}, charts a new course by decoupling feature extraction from reasoning, introducing a pre-trained LLM as a dedicated reasoning engine to address this limitation.

\noindent\textbf{Vision-Language Fusion for LLMs.}
Large Language Models (LLMs) are massive, transformer-based networks pre-trained on text corpora, demonstrating emergent capabilities in complex reasoning that form the foundation for modern AI~\cite{ref:gpt3}. A significant research frontier is extending their reasoning to multi-modal contexts like vision. This adaptation is made feasible by parameter-efficient fine-tuning (PEFT) techniques such as Low-Rank Adaptation (LoRA)~\cite{ref:lora}, which we employ. A primary bottleneck remains the effective translation of rich visual information into a format LLMs can process, as simply projecting a final global feature often discards critical details. Our proposed \textbf{enhanced fusion projector (EFP)} carves a distinct niche. By using an MLP to learn a non-linear mapping from concatenated multi-level visual features to a single, information-dense token, the EFP creates the high-quality representation necessary for robust LLM-based reasoning.

\vspace{-0.75em}
\section{Methodology}
\begin{figure}[t!]
\centering
\includegraphics[width=0.9\textwidth]{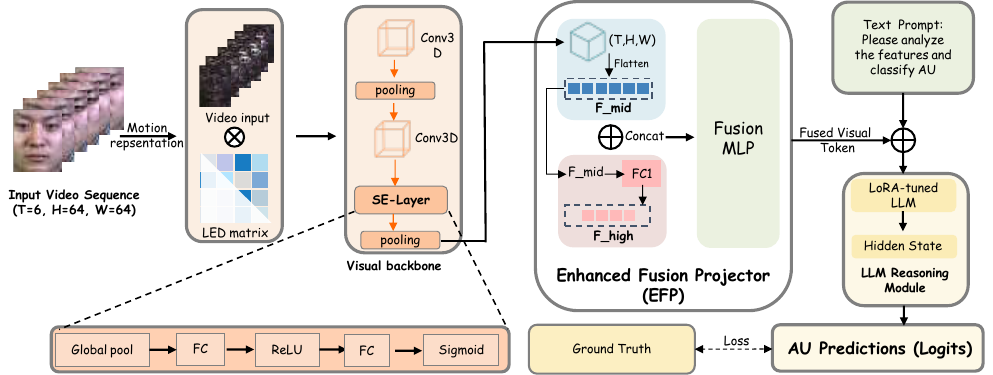}
\vspace{-0.6em}
\caption{The overall framework of our proposed AU-LLM. A video sequence is first processed by the visual backbone, which uses a LED matrix~\cite{ref:led} and 3D-CNNs to extract multi-level features ($\mathbf{F_{\text{mid}}}$ and $\mathbf{F_{\text{high}}}$). The Enhanced Fusion Projector (EFP) then fuses these features via concatenation and an MLP into a single, information-dense visual token ($\mathbf{T_v}$). This visual token is combined with a text prompt and fed to a LoRA-tuned LLM for reasoning and final AU classification.}
\vspace{-1.2em}
\label{fig:framework}
\end{figure}

\vspace{-0.75em}
In this section, we present our proposed framework, \textbf{AU-LLM}, designed for enhancing micro-expression facial Action Unit detection by leveraging the reasoning capabilities of Large Language Models (LLMs). As illustrated in Figure~\ref{fig:framework}, the architecture comprises a visual backbone for multi-level feature extraction, our novel Enhanced Fusion Projector (EFP), an LLM reasoning module, and a final classification head.

\vspace{-0.75em}
\subsection{Visual Backbone}
\vspace{-0.75em}
The visual backbone is engineered to extract a rich set of spatio-temporal features from an input video sequence $\mathbf{V} \in \mathbb{R}^{T \times H \times W}$, where $T=6$ is the number of frames, and $H=W=64$ are the spatial dimensions.

\textbf{Temporal Filtering.} To amplify subtle motion cues crucial for micro-expression AUs, we first apply a custom temporal filtering operation, termed the Laplacian of Exponential of Difference (LED) module~\cite{ref:led}. This module re-weights the temporal sequence by applying a learnable filter matrix $\mathbf{W} \in \mathbb{R}^{T \times T}$, computed as:
\begin{equation}
\begin{aligned}
W_{i,j} = \begin{cases} 
      \alpha \left( (1-r_1)^{j-i} r_1^{\min(1,i)} - (1-r_2)^{j-i} r_2^{\min(1,i)} \right) & \text{if } j > i ,\\
      \alpha (r_1 - r_2) & \text{if } j = i, \\
      0 & \text{if } j < i,
   \end{cases}
\end{aligned}
\end{equation}
where $\alpha, r_1, r_2$ are learnable parameters initialized based on prior work to capture onset-apex-offset patterns~\cite{ref:led}. The filtered video representation $\mathbf{V}'$ is obtained via a normalized matrix multiplication, enhancing transient changes.

\textbf{Spatio-Temporal Feature Extraction.} The temporally enhanced sequence $\mathbf{V}'$ is processed by a 3D-CNN. The network consists of 3D convolutional layers, batch normalization, and dropout. A Squeeze-and-Excitation (SE) Layer is integrated after the second convolution to perform channel-wise feature recalibration, allowing the network to focus on more informative channels for AU detection~\cite{ref:se}. We extract features from two distinct stages of this backbone:
\begin{itemize}[leftmargin=*]
    \item \textbf{Mid-level Features ($\mathbf{F_{\text{mid}}}$)}: Extracted after the second 3D convolutional block and pooling layer, these features retain significant spatial information and describe local textures and shapes~\cite{ref:physllm}.
    \item \textbf{High-level Features ($\mathbf{F_{\text{high}}}$)}: Extracted after a fully-connected layer that processes the flattened mid-level features, these features are more abstract and capture global semantic information about the facial region.
\end{itemize}

\vspace{-1.5em}
\subsection{Enhanced Fusion Projector (EFP)}
\vspace{-0.5em}
A critical challenge in leveraging LLMs for visual tasks is effectively translating rich, high-dimensional visual data into the compact, discrete token space that LLMs operate on. Simply projecting final-layer features can discard vital mid-level details crucial for fine-grained tasks like AU detection. To address this, we designed the Enhanced Fusion Projector (EFP). It intelligently fuses the multi-level visual features into a single, information-dense token for the LLM. The mid-level and high-level feature vectors are first flattened and concatenated:
\begin{equation}
    \mathbf{f_{cat}} = \text{Concat}(\text{Flatten}(\mathbf{F_{\text{mid}}}), \mathbf{F_{\text{high}}}).
\end{equation}

This vector $\mathbf{f_{cat}}$, containing a comprehensive visual summary, is then passed through a dedicated fusion MLP. The MLP learns a powerful non-linear transformation to both fuse the features and project them into the LLM's embedding space:
\begin{equation}
    \mathbf{T_{v}} = \sigma(\mathbf{W}_2(\text{ReLU}(\mathbf{W}_1 \mathbf{f_{cat}} + \mathbf{b}_1)) + \mathbf{b}_2),
\end{equation}
where $\mathbf{W}_1, \mathbf{b}_1, \mathbf{W}_2, \mathbf{b}_2$ are the weights and biases of the MLP, and $\sigma$ is a non-linear activation. The resulting visual token, $\mathbf{T_{v}}$, is a highly distilled vector that encapsulates both the local textural details (from $\mathbf{F_{\text{mid}}}$) and the global semantic context (from $\mathbf{F_{\text{high}}}$) of the facial movement. In the next step, this single token acts as a soft prompt or a visual instruction for the LLM. It is prepended to a task-specific text prompt, and the combined sequence of embeddings is fed into the LLM to perform the core reasoning task. This learned fusion is superior to a simple linear projection, as the MLP can model complex interactions between mid- and high-level features, distilling the most salient AU-related information for the final classification.

\vspace{-1.25em}
\subsection{LLM Reasoning Module}
\vspace{-0.5em}
To steer the LLM's powerful reasoning capabilities towards our specific task, we must provide it with both the visual evidence and the analytical context. This is achieved by prepending the visual token $\mathbf{T_{v}}$, which serves as a soft prompt encapsulating the visual information, to a task-specific text prompt (e.g. Analyze the facial features to classify action units:). This fusion of visual evidence and textual instruction creates a multi-modal input sequence that is subsequently fed into a pre-trained LLM. We efficiently fine-tune the LLM using Low-Rank Adaptation (LoRA), which introduces low-rank matrices into the query and value projections of the self-attention layers for efficient adaptation. Finally, the hidden state of the last token from the LLM, which encapsulates the model's unified reasoning over both visual and textual cues, is passed to a linear classifier to produce the final logits $\mathbf{z} \in \mathbb{R}^{N_{AUs}}$ for each AU.

\vspace{-1.25em}
\subsection{Loss Function and Optimization}
\vspace{-0.5em}
To address the severe class imbalance inherent in multi-label AU detection, we employ the Asymmetric Loss (ASL)~\cite{ref:asl}. ASL is designed to mitigate the dominance of negative samples by applying different focusing parameters to positive and negative examples. The loss is formulated as:
\begin{equation}
    \mathcal{L}_{\text{ASL}} = -\frac{1}{N} \sum_{i=1}^{N} \sum_{j=1}^{B} \left[ y_{ij} (1-p_{ij})^{\gamma_{+}} \log(p_{ij}) + (1-y_{ij}) (p_{m, ij})^{\gamma_{-}} \log(1-p_{m, ij}) \right],
\end{equation}
where $p_{ij}$ is the predicted probability and $y_{ij}$ is the ground truth label. The focusing parameters $\gamma_{+}$ and $\gamma_{-}$ allow for flexible weighting of samples~\cite{ref:AUformer}. Based on our implementation, we set $\gamma_{+}=0$ and $\gamma_{-}=4$.

\vspace{-1.4em}
\section{Experiments}
\vspace{-0.6em}
\subsection{Datasets and Protocol}
\vspace{-0.5em}
We validate our approach on two publicly available spontaneous micro-expression datasets, both of which are standard benchmarks in the field.
\textbf{CASME II}~\cite{ref:casme2_ref} contains 247 micro-expression samples from 26 subjects. Captured with a high-speed camera (200 fps) under controlled laboratory conditions, it is one of the most widely used datasets for micro-expression analysis. Following standard protocol, we detect 8 AUs: AU1, AU2, AU4, AU7, AU12, AU14, AU15, and AU17.
\textbf{SAMM}~\cite{ref:samm_ref} consists of 159 micro-expression samples from 32 subjects with diverse ethnic backgrounds. This dataset also utilizes a high-speed camera and provides a challenging testbed for evaluating model generalization. We conduct cross-dataset validation on SAMM, detecting 4 AUs common to its annotation: AU2, AU4, AU7, and AU12.

\begin{table}[t]
\centering
\vspace{-1.9em}
\caption{Micro-Expression AU Detection Performance Comparison (CASME II). F1-scores (\%) are reported. Best results are in \textbf{bold}.}
\label{tab:au_comparison_casme2}
\resizebox{0.85\textwidth}{!}{
\begin{tabular}{l|cccccccc|c}
\toprule
\textbf{Method} & \textbf{AU1} & \textbf{AU2} & \textbf{AU4} & \textbf{AU7} & \textbf{AU12} & \textbf{AU14} & \textbf{AU15} & \textbf{AU17} & \textbf{Avg.} \\
\midrule
LBPTOP~\cite{ref:lbptop} & 80.9 & 60.7 &\textbf{ 89.8} & 56.2 & 69.3 & 63.2 & 52.2 & 75.8 & 68.5 \\
Resnet18~\cite{ref:resnet} & 58.9 & 76.2 & 82.7 & 48.7 & 59.3 & 64.2 & 48.1 & 62.9 & 62.6 \\
Resnet34~\cite{ref:resnet} & 55.1 & 69.2 & 83.2 & 55.4 & 62.8 & 56.7 & 52.2 & 62.9 & 62.2 \\
Fitnet~\cite{ref:fitnet} & 65.5 & 74.3 & 79.1 & 46.8 & 61.7 & 62.4 & 53.9 & 57.6 & 62.7 \\
SP~\cite{ref:sp} & 69.0 & 68.5 & 77.7 & 51.1 & 57.0 & 65.3 & 56.6 & 61.3 & 63.3 \\
AT~\cite{ref:at} & 65.0 & 62.7 & 83.1 & 57.6 & 58.6 & 62.5 & 46.6 & 65.6 & 62.7 \\
SCA~\cite{ref:sca} & 64.1 & 76.7 & 81.1 & 56.2 & 61.3 & 61.9 & 68.8 & 64.4 & 66.8 \\
    DVASP~\cite{ref:dvasp} & 72.6 & 72.1 & \textbf{89.8} & 56.9 & \textbf{79.6} & 68.5 & 71.5 & 70.0 & 72.6 \\
Resnet18 LED~\cite{ref:led} & 85.8 & 81.7 & 89.0 & 53.9 & 71.4 & 78.5 & 74.5 & 83.5 & 77.3 \\
SSSNet LED~\cite{ref:led} & \textbf{92.6} & 83.7 & 88.6 & 63.7 & 76.6 & 74.4 & 71.5 & 76.1 & 78.4 \\
\midrule
AU-DeepSeek R1(1.5B) & 92.2 & \textbf{87.0} & {89.5} & \textbf{68.1} & {79.0} & 75.1 & \textbf{79.0} & 81.0 & \textbf{81.4} \\
AU-Qwen2(1.5B) & 88.4 & 82.1 & {89.5} & 65.7 & 75.8 & 74.7 & 71.7 & 82.7 & 78.8 \\
AU-Qwen2.5(1.5B) & 89.1 & 86.6 & {89.5} & 65.2 & 74.4 & \textbf{77.9} & 67.6 & \textbf{85.8} & 79.5 \\
\bottomrule
\end{tabular}}
\vspace{-2.0em}
\end{table}

For fair and robust evaluation, we employ the stringent Leave-One-Subject-Out (LOSO) cross-validation protocol~\cite{ref:led} to ensure subject-independent evaluation. The primary metric is the macro F1-score, which is well-suited for AU detection tasks with significant class imbalance as it treats each AU class equally~\cite{ref:varanka2023data}. The F1-score is the harmonic mean of precision and recall, calculated for each action unit class $c$ as:
\begin{equation}
    \mathcal{F}_{1,c} = 2 \cdot \frac{\text{Precision}_c \cdot \text{Recall}_c}{\text{Precision}_c + \text{Recall}_c}.
\end{equation}
The final reported score is the unweighted average of these individual F1-scores across all AU classes.

\vspace{-1.25em}
\subsection{Implementation Details}
\vspace{-0.5em}
Our framework, AU-LLM, is implemented in PyTorch. The visual backbone is trained from scratch with random initialization for each fold of the cross-validation. We use the Adam optimizer with a learning rate of $3 \times 10^{-5}$ and a weight decay of $0.005$. The model is trained with a batchsize of 256. We utilize several 1.5B-parameter LLMs, including \textbf{Qwen2-1.5B}, \textbf{Qwen2.5-1.5B}, and \textbf{DeepSeek-R1-Distill-Qwen-1.5B}, loaded from Hugging Face~\cite{ref:deepseek_r1}~\cite{ref:qwen2}. Parameter-Efficient Fine-Tuning is performed using Low-Rank Adaptation (LoRA) with a rank $r=16$ and alpha $\alpha=32$ applied to the query and value matrices of the attention blocks. All experiments are run on a single NVIDIA H100 GPU.

\vspace{-1.25em}
\subsection{Comparison with State-of-the-Art Methods}
\vspace{-0.5em}
We compare AU-LLM with several state-of-the-art (SOTA) methods on both datasets. As shown in Table \ref{tab:au_comparison_casme2}, our model variants achieve superior performance on CASME II. Notably, \textbf{AU-deepseek R1(1.5B)} achieves a mean F1-score of \textbf{81.4\%}, surpassing the previous best method SSSNet LED (78.4\%) by a significant margin of 3.0\%. Our method shows particularly strong performance on key AUs like AU2 (87.0\%), AU4 (89.5\%), and AU7 (68.1\%), demonstrating its effectiveness in capturing diverse facial muscle movements.

The results on the SAMM dataset, shown in Table \ref{tab:sota_comparison_samm}, further validate the generalization capability of our framework. In this cross-dataset setting, \textbf{AU-deepseek R1(1.5B)} again achieves the highest average F1-score of \textbf{61.9\%}, outperforming all previous methods. This robust performance highlights the benefit of our EFP module in creating a rich, dataset-agnostic visual representation that enables the LLM to reason effectively even on unseen data distributions.

\begin{table}[t]
\centering
\vspace{-1.9em}

\caption{Micro-Expression AU Detection Performance Comparison (SAMM). F1-scores (\%) are reported. Best results are in \textbf{bold}.}
\label{tab:sota_comparison_samm}
\resizebox{0.7\textwidth}{!}{
\begin{tabular}{l|cccc|c}
\toprule
\textbf{Method} & \textbf{AU2} & \textbf{AU4} & \textbf{AU7} & \textbf{AU12} & \textbf{Avg.} \\
\midrule
LBP-TOP~\cite{ref:lbptop} & 58.8 & 47.9 & 44.5 & 49.5 & 50.2 \\
ResNet18~\cite{ref:resnet} & 49.7 & 49.1 & 46.1 & 40.5 & 46.4 \\
ResNet34~\cite{ref:resnet} & 44.0 & 55.2 & 38.0 & 40.5 & 44.4 \\
Fit-18~\cite{ref:fitnet} & 54.1 & 51.2 & 44.5 & 48.3 & 49.5 \\
SP-18~\cite{ref:sp} & 42.8 & 64.2 & 38.1 & 49.5 & 48.7 \\
AT-18~\cite{ref:at} & 47.2 & 60.5 & 43.5 & 38.0 & 47.3 \\
SCA~\cite{ref:sca} & 45.7 & 59.2 & 43.9 & 53.2 & 50.5 \\
DVASP-18~\cite{ref:dvasp} & 47.8 & 67.5 & 48.1 & 44.7 & 52.0 \\
Resnet18 LED~\cite{ref:led} & 57.4 & 71.3 & 53.2 & 47.3 & 57.3 \\
SSSNet LED~\cite{ref:led} & 61.5 & 62.4 & 45.2 & 47.8 & 54.2 \\
\midrule
AU-DeepSeek R1(1.5B) & \textbf{66.9} & \textbf{71.9} & \textbf{55.8} & 52.8 & \textbf{61.9} \\
AU-Qwen2(1.5B) & 63.6 & 65.2 & 49.2 & 53.7 & 57.9 \\
AU-Qwen2.5(1.5B) & 63.2 & 67.6 & 50.7 & \textbf{57.6} & 59.7 \\
\bottomrule
\end{tabular}}
\vspace{-0.2em}
\end{table}

\begin{table}[t]
\centering
\vspace{-1.9em}
\caption{Ablation study on CASME II and SAMM datasets. Average F1-scores (\%) are reported. Model names are abbreviated. Best results are in \textbf{bold}.}
\label{tab:ablation}
\vspace{-1.2em}
\begin{minipage}{.48\textwidth}
  \centering
  \subcaption{CASME II}
  \vspace{-0.5em}
  \begin{tabular}{l|ccc}
    \toprule
    \textbf{Variant} & \textbf{D.R1} & \textbf{Q2} & \textbf{Q2.5} \\
    \midrule
    $F_{\text{high}}$ & 80.1 & 79.8 & 78.8 \\
    $F_{\text{mid}}$ & 79.3 & 78.7 & 78.7 \\
    EFP-- & 78.6 & 77.8 & 77.7 \\
    Prompt Learning&80.4  & 78.6& 79.3 \\
    all & \textbf{81.4} & \textbf{78.8} & \textbf{79.5} \\
    \bottomrule
  \end{tabular}
\end{minipage}%
\hfill
\begin{minipage}{.48\textwidth}
  \centering
  \subcaption{SAMM}
  \vspace{-0.5em}
  \begin{tabular}{l|ccc}
    \toprule
    \textbf{Variant} & \textbf{D.R1} & \textbf{Q2} & \textbf{Q2.5} \\
    \midrule
    $F_{\text{high}}$ & 54.5 & 55.5 & 54.3 \\
    $F_{\text{mid}}$ & 54.7 & 55.6 & 55.5 \\
    EFP-- & 58.2 & 56.3 & 55.3 \\
    Prompt Learning& 59.6 & 56.4 &57.6  \\
    all & \textbf{61.9} & \textbf{57.9} & \textbf{59.7} \\
    \bottomrule
  \end{tabular}
\end{minipage}
\vspace{-1.5em}
\end{table}

\vspace{-1.25em}
\subsection{Ablation Study}
\begin{figure}[t]
\centering
\begin{subfigure}{0.49\textwidth}
    \centering
    \includegraphics[width=0.8\linewidth]{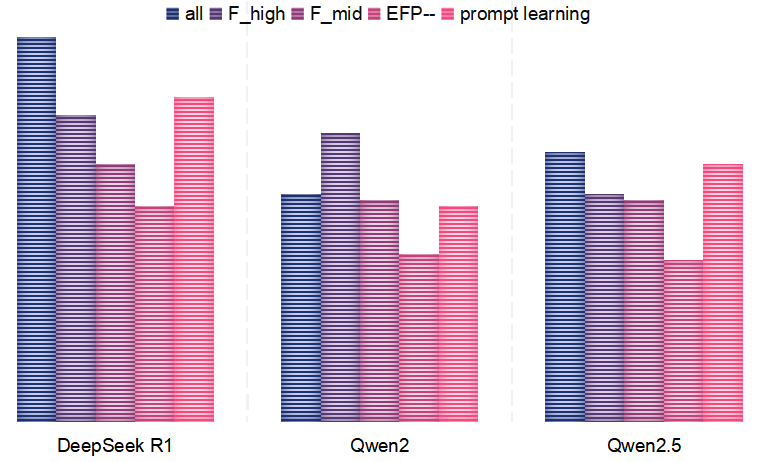}
    \vspace{-0.5em}
    \caption{Ablation results on CASME II.}
    \label{fig:ablation_casme2}
\end{subfigure}
\hfill
\begin{subfigure}{0.49\textwidth}
    \centering
    \includegraphics[width=0.8\linewidth]{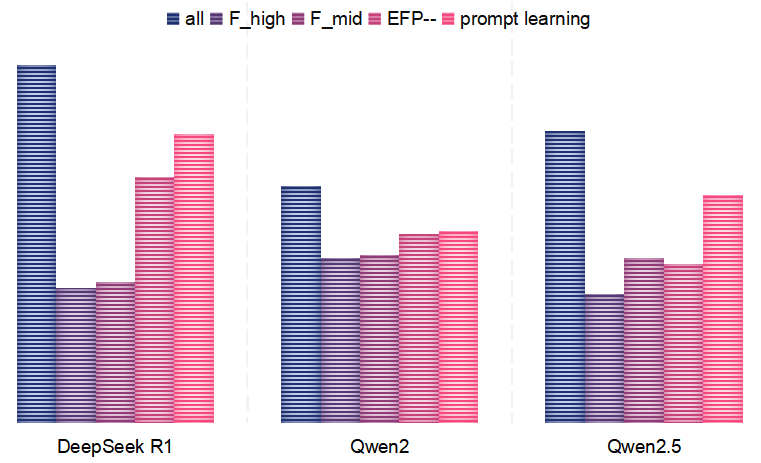}
    \vspace{-0.5em}
    \caption{Ablation results on SAMM.}
    \label{fig:ablation_samm}
\end{subfigure}
\vspace{-0.8em}
\caption{Visualization of ablation studies on both datasets, showing the performance of the full model (all) versus variants lacking certain components.}
\label{fig:ablation_plots}
\vspace{-1.6em}
\end{figure}

\vspace{-0.5em}
\textbf{Ablation on Modules.} To validate our key design choices, we conducted ablation studies on the CASME II and SAMM datasets. We designed three variants to isolate the contributions of our core components: (1) using only high-level features (\textbf{$\mathbf{F}_{\text{high}}$}) or (2) only mid-level features (\textbf{$\mathbf{F}_{\text{mid}}$}) to test the necessity of fusing multi-level information, and (3) \textbf{EFP--}, which replaces the EFP's MLP with a simple linear layer, to verify the benefit of non-linear fusion. As shown in Table \ref{tab:ablation}, the full model (all) consistently outperforms all ablated versions across both datasets. The performance drop when using single-level features confirms that both local and global cues are vital for capturing the full spectrum of AU-related movements. The superiority over the EFP-- variant highlights the importance of the MLP's non-linear fusion capability. These results are complemented by the visualizations in Figure \ref{fig:ablation_plots}, which provide a graphical representation of the performance differences.

\noindent\textbf{Ablation on Prompting Strategy.}
To further validate our dynamic visual prompting strategy, we introduce an ablation study that replaces our EFP module with a strong baseline: \textbf{Learnable Text Prompts}~\cite{ref:coop}. This variant uses a set of static, learnable embedding vectors optimized to act as a general textual instruction for the AU detection task, instead of generating a dynamic visual token for each sample. This experiment directly contrasts our "early fusion" approach, where rich visual information is integrated before reaching the LLM, with a "late fusion" approach. The expected superior performance of our EFP-based model would strongly demonstrate that for fine-grained visual tasks like micro-expression detection, dynamically generating an instance-specific visual prompt is a more effective strategy than using a static, task-level textual prompt, thus validating the design of our EFP module.

\begin{table}[t]
\caption{F1-scores (in \%) for cross-domain evaluations between CASME II and SAMM. }
\centering
\resizebox{0.75\textwidth}{!}{
\begin{tabular}{l|cccc|c|cccc|c}
\toprule
Datasets& \multicolumn{5}{c|}{CASME II $\rightarrow$ SAMM} & \multicolumn{5}{c}{SAMM $\rightarrow$ CASME II} \\
\midrule
AU & 2 & 4 & 7 & 12 & \textbf{Avg.}  & 2 & 4 & 7 & 12 & \textbf{Avg.} \\
\midrule
ResNet-18~\cite{ref:resnet} & \textbf{47.0} & 15.7 & \textbf{41.5} & 45.1 & 37.3 & 47.8 & \textbf{47.4} & {47.9} & 46.4 &{47.4} \\
LED SSSNet~\cite{ref:led} & \textbf{47.0} & 12.6 & \textbf{41.5} & 45.1 & 36.5 & 53.2 & 44.1 & 40.2 & 46.5 & 46.1 \\
\textbf{AU-LLM (Ours)} & \textbf{47.0} & \textbf{62.4} &\textbf{41.5} & \textbf{48.8} &\textbf{49.9} & \textbf{67.8} & {39.4} & \textbf{50.4} & \textbf{53.6} & \textbf{52.8}  \\
\bottomrule
\end{tabular}}
\label{tab:CrossDomain}
\vspace{-0.8em}
\end{table}

\vspace{-1.25em}
\subsection{Cross-Domain Evaluation}
\vspace{-0.5em}
To further assess the model's generalization capability, we conducted bidirectional cross-domain evaluations between the CASME II and SAMM datasets on their shared AUs. As presented in Table \ref{tab:CrossDomain}, our model demonstrates superior generalization. Under the \textbf{CASME II $\rightarrow$ SAMM} protocol (training on CASME II, evaluating on SAMM), our AU-LLM achieves an average F1-score of 49.9\%, significantly outperforming the baseline LED SSSNet (36.5\%) by \textbf{13.4} percentage points. Conversely, under the \textbf{SAMM $\rightarrow$ CASME II} protocol, our model achieves 52.8\%, surpassing the baseline (46.1\%) by \textbf{6.7} percentage points. These results strongly validate the superior generalization ability of our method.

\vspace{-1.25em}
\subsection{Visualization and Analysis}
\begin{figure}[t]
\centering
\includegraphics[width=0.75\textwidth]{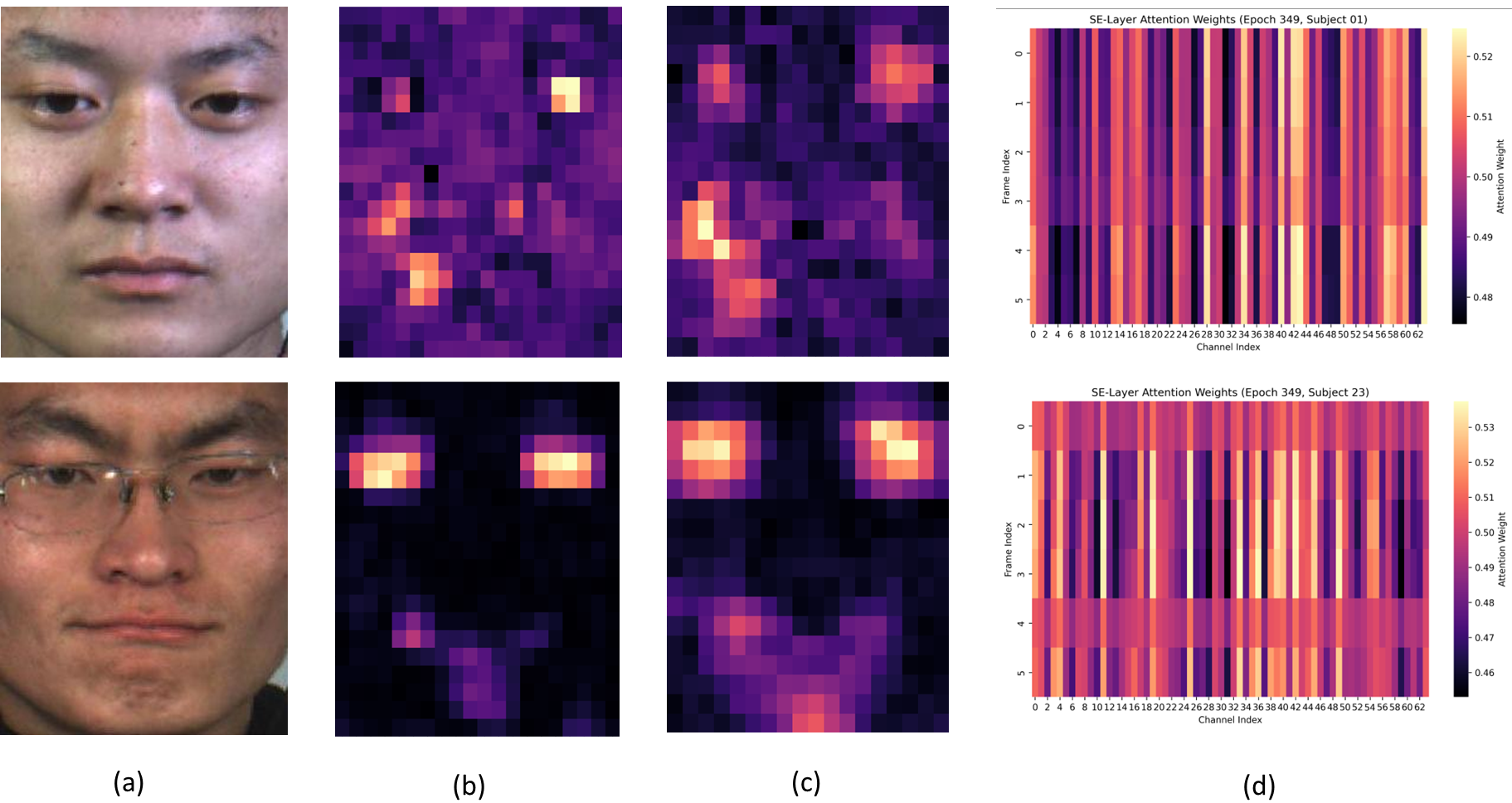} 
\vspace{-0.8em}
\caption{Visualization of model attention. (a) Original micro-expression samples. (b) Heatmaps from a baseline 3D-CNN. (c) Heatmaps from our visual backbone, showing more precise focus. (d) SE-Layer channel attention weights over time. \textbf{Top row:} The ground truth is AU12 (Lip Corner Puller). \textbf{Bottom row:} The ground truth is a combination of AU4 (Brow Lowerer), AU15 (Lip Corner Depressor), and AU17 (Chin Raiser).}
\vspace{-1.8em}
\label{fig:visualize}
\end{figure}

\vspace{-0.5em}
To intuitively understand how our model's visual backbone processes facial information, we visualize the feature heatmaps and compare them against a baseline 3D-CNN. As shown in Figure \ref{fig:visualize}, our model demonstrates a superior ability to focus on the correct AU-related facial regions.

The qualitative analysis in Figure \ref{fig:visualize} provides strong evidence for our model's effectiveness. For both a simple case (AU12, top row) and a complex combination (AU4+15+17, bottom row), our model's heatmaps (c) accurately localize the corresponding facial regions. In contrast, the baseline model (b) shows diffuse or incorrect attention. Furthermore, the SE-Layer's channel attention maps (d) illustrate the model's ability to dynamically re-weight feature channels to amplify the most informative signals. This confirms that our visual backbone learns a more precise and interpretable feature representation, providing high-quality input for the subsequent LLM reasoning.

\vspace{-1.0em}
\section{Conclusion}
\vspace{-0.7em}
In this paper, we introduced \textbf{AU-LLM}, a framework that, for the first time, successfully leverages Large Language Models for the challenging task of micro-expression Action Unit detection. Our innovation, the \textbf{Enhanced Fusion Projector (EFP)}, effectively bridges the vision-language semantic gap by fusing multi-level visual features into a compact, information-rich token, empowering the LLM to perform nuanced reasoning over subtle facial cues. This work validates the significant potential of applying advanced reasoning engines to this fine-grained domain. For future work, our focus will shift towards fully harnessing the capabilities of LLMs. We aim to build upon this foundation to develop systems capable of more complex micro-expression reasoning and interactive, context-aware question-answering, moving beyond simple classification to a deeper understanding of concealed emotions.

\bibliographystyle{splncs04}
\bibliography{references}

\end{document}